\documentclass[letterpaper, 10 pt, conference]{ieeeconf}  %

\IEEEoverridecommandlockouts                              %

\usepackage{amsmath,amssymb,amsfonts}
\usepackage{svg}
\usepackage{multirow}
\usepackage[caption=false]{subfig}

\definecolor{darkgreen}{rgb}{0.09, 0.32, 0.24}

\title{\LARGE \bf 
UniCal: a Single-Branch Transformer-Based Model for Camera-to-LiDAR Calibration and Validation
}

\author{
    Mathieu Cocheteux$^{1\dag}$, Aaron Low$^{2\dag}$, Marius Brühlmeier$^{1}$%
    \thanks{$^{1}$Mathieu Cocheteux {\tt\small mathieu@cocheteux.eu} and Marius Brühlmeier were with Motional at the time of this work.}%
    \thanks{$^{2}$Aaron Low {\tt\small aaron.low@motional.com} is with Motional}%
    \thanks{$^{\dag}$Contributed to the work equally}
}

\begin{document}

\maketitle
\thispagestyle{empty}
\pagestyle{empty}

\begin{abstract}
We introduce a novel architecture, UniCal, for Camera-to-LiDAR (C2L) extrinsic calibration which leverages self-attention mechanisms through a Transformer-based backbone network to infer the 6-degree of freedom (DoF) relative transformation between the sensors. Unlike previous methods, UniCal performs an early fusion of the input camera and LiDAR data by aggregating camera image channels and LiDAR mappings into a multi-channel unified representation before extracting their features jointly with a single-branch architecture. This single-branch architecture makes UniCal lightweight, which is desirable in applications with restrained resources such as autonomous driving. Through experiments, we show that UniCal achieves state-of-the-art results compared to existing methods. We also show that through transfer learning, weights learned on the calibration task can be applied to a calibration validation task without re-training the backbone.
\end{abstract}

\section{Introduction}
Autonomous vehicles, as well as other robotic systems, rely on sensors to perceive their environment. Most systems use a variety of sensors, including cameras, LiDARs, as well as radars. Fusing data from multiple sensors is a common practice to leverage the advantages of each sensor. This requires having the precise value of their extrinsic calibration, which is a transformation in the Lie group SE(3) represented by a 6-degree of freedom (DoF) transformation consisting of the relative translation and rotation between the two sensor poses. Camera-to-LiDAR (C2L) calibration is one of the most common sensor fusion strategies as it is valuable to combine the visual information obtained by cameras with the spatial and occupancy information obtained by LiDARs. With this information, downstream tasks such as mapping, localization, and planning can be carried out. 

Traditionally, C2L calibration is performed with offline methods~\cite{zhangExtrinsicCalibrationCamera2004a}\cite{rodriguezf.ExtrinsicCalibrationMultilayer2008}\cite{geigerAutomaticCameraRange2012}. This usually involves pre-collecting a series of frame in a controlled environment to carry out iterative optimization. However, this does not consider the scenario where sensor position changes during the normal operation of a robot, by weather conditions, mechanical vibrations, or collision. This implies the need for online methods that can perform calibration in real-time. Existing solutions have already been introduced to tackle this problem~\cite{nagySFMSemanticInformation2019}. These solutions tend to rely on geometric feature detection and optimization as in~\cite{yuanPixellevelExtrinsicSelf2021}. Deep learning methods were first introduced with RegNet~\cite{schneiderRegNetMultimodalSensor2017} which proposed a dual-branch architecture that processed the camera and LiDAR features separately before passing the features through a matching layer before regressing the output transformation. Following~\cite{schneiderRegNetMultimodalSensor2017} various methods build upon a dual-branch architecture and manage to improve upon the results~\cite{lvLCCNetLiDARCamera2021}\cite{iyerCalibNetGeometricallySupervised2018}. For the deep learning methods, experiments focus on scenes collected by a single camera located at the front of the vehicle.

Our work, UniCal, differs by its novel architecture that can be used as a baseline for future works. Unlike the dual-branch architecture used in other works~\cite{zhangExtrinsicCalibrationCamera2004a}\cite{rodriguezf.ExtrinsicCalibrationMultilayer2008}\cite{geigerAutomaticCameraRange2012}, it merges the raw inputs directly and relies on a single backbone to perform the task. Its single-branch architecture enables our network to be more lightweight. It is also extensible as it is not required to build a new feature encoder should we want to modify the input. This architecture is also the first to use a Transformer-based backbone, MobileViT~\cite{mehtaMobileViTLightweightGeneralpurpose2022} to leverage self-attention mechanisms and identify areas with the most significant features. Similar to previous works~\cite{schneiderRegNetMultimodalSensor2017}\cite{lvLCCNetLiDARCamera2021}\cite{iyerCalibNetGeometricallySupervised2018}\cite{wuNetCalibNovelApproach2021}, we evaluate on the Kitti~\cite{geigerVisionMeetsRobotics2013} dataset. We also evaluate UniCal on our own dataset collected from multiple autonomous vehicle logs which we refer to as Motional drivelog data which we elaborate further in Section \ref{motional_data_sss}. Unlike Kitti, this dataset consists of varying vehicles and features a suite of cameras surrounding the vehicle which provides various points of view.  From our experiments, we achieved on par or better compared to the current state-of-the-art for both datasets.

\begin{figure}[t]
    \centering
    \includeinkscape[width=\columnwidth]{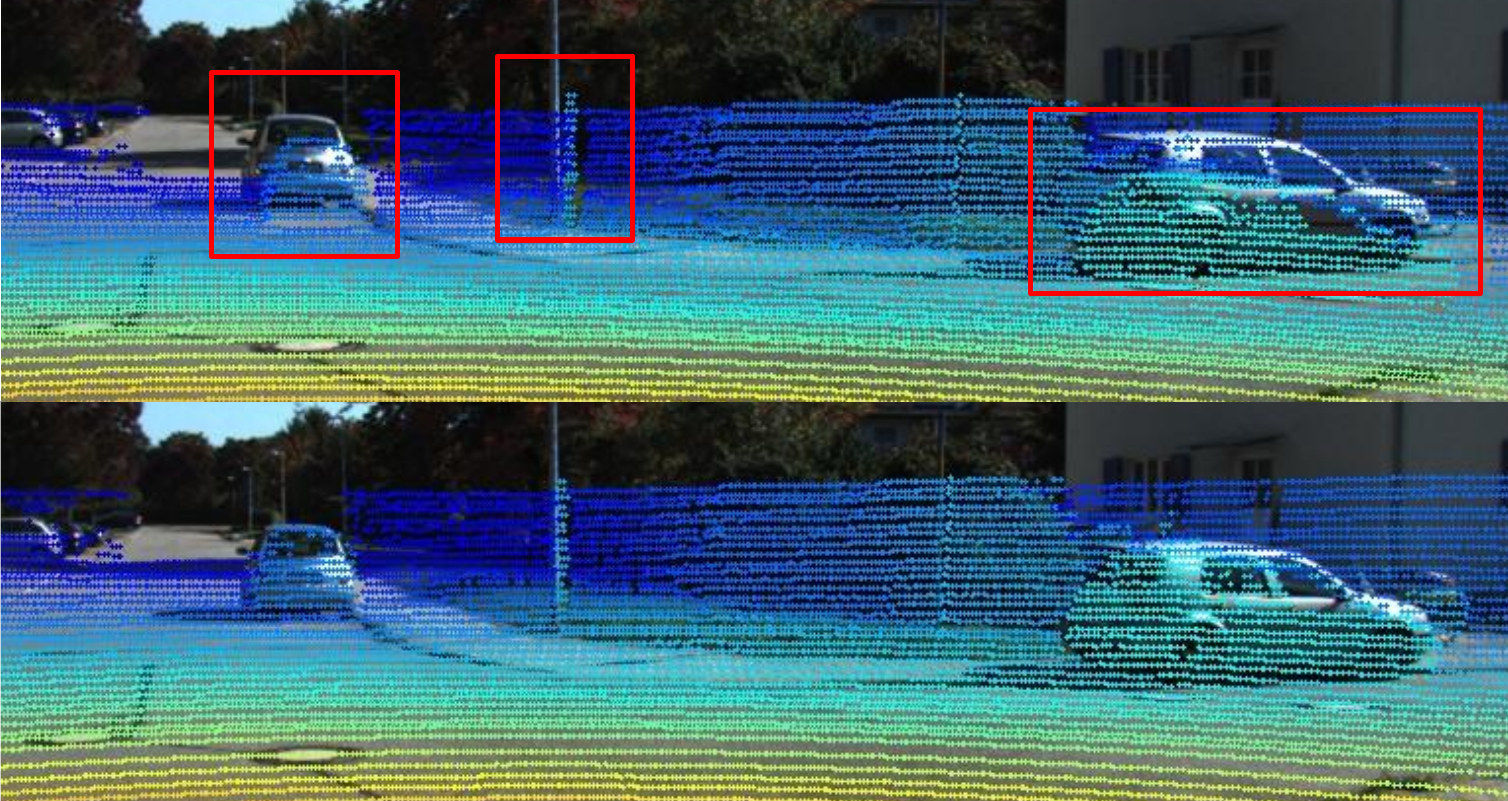}
    \caption{Projection of LiDAR points with color-mapped depth onto the corresponding camera image. Frame from the Kitti~\cite{geigerVisionMeetsRobotics2013} raw dataset. (\textbf{Top}) De-calibrated on rotation and translation parameters. (\textbf{Bottom}) Re-calibrated with UniCal.}
    \label{fig:c2l}
\end{figure}

Furthermore, we extend UniCal to perform the task of C2L calibration validation. The task is to determine the validity of a given set of calibration parameters for a pair of sensors. This is relevant for autonomous vehicle use cases as it can be used to determine instantly if a vehicle is well-calibrated before allowing it to operate. This problem can be reduced to a binary classification task (calibrated/de-calibrated), where the calibrated class is defined as having a de-calibration smaller than a chosen sensitivity margin. We determined by experimentation that training UniCal for a calibration task and then using transfer learning with a classification head to solve this issue delivered better results than directly learning this task with UniCal architecture and outperforming existing calibration architectures.

To summarize, our main contributions are:
\begin{itemize}
    \item A lightweight single-branch architecture which has up to 3 to 10 times less parameters than the state-of-the-art architectures;
    \item Our model achieves up to 7 times lower mean average error compared to RegNet~\cite{schneiderRegNetMultimodalSensor2017};
    \item An early fusion of all inputs in a unified representation;
    \item Leveraging Transformers and self-attention to learn meaningful features from unstructured environments;
    \item A transfer learning technique that improves camera-to-LiDAR calibration validation by training a classification head on top of frozen calibration network weights, achieving $98\%$ accuracy.
\end{itemize}

\section{Related Works}
Sensor calibration is a well-defined and broad research topic including diverse tasks. We decided to focus our work on camera-to-LiDAR (C2L) extrinsic calibration. This section will give a brief overview of the field.\\

\subsubsection{Target-based methods} This task has most often been solved with offline, target-based methods.~\cite{zhangExtrinsicCalibrationCamera2004a} proposes to use a common checkerboard target, which can be seen both as a 3D object by the LiDAR and as a pattern by the camera. Other works explore different target shapes. For example,~\cite{rodriguezf.ExtrinsicCalibrationMultilayer2008} proposes a target with a circle and a hole to improve visibility by sensors in outdoor scenes. Finally,~\cite{geigerAutomaticCameraRange2012} introduces an automated calibration process that uses multiple checkerboard targets in an indoor setting to converge toward a solution with one shot in under one minute. Traditional target-based methods are accurate but require specific equipment and environment. Moreover, they require users to gain experience for optimal target positioning. These methods require tuning a lot
of parameters, which introduces heuristics and requires
iterative optimization which takes time to converge. These methods can thus be slow and costly to use regularly on a vehicle.

\subsubsection{Targetless methods} More recent works have achieved automated, targetless methods. ~\cite{nagySFMSemanticInformation2019} proposes to use structure from motion and semantic information to treat the task as a point cloud registration problem.~\cite{yuanPixellevelExtrinsicSelf2021} explores detecting natural edge features in both modalities to align them. These methods are automated, targetless, and can be computed online. However, they still rely on computation-heavy optimization and require a feature rich environment to provide a correct distribution of the selected features.

\subsubsection{Machine Learning methods} Considering the recent success of neural networks in performing computer vision tasks, some works explored their use to perform camera-to-LiDAR calibration. This was first introduced by RegNet~\cite{schneiderRegNetMultimodalSensor2017}, a seminal work that proposed numerous ideas that have since then been commonly reused in other works:
\begin{itemize}
    
\item  A training method that consists in generating artificially de-calibrated samples from the ground truth;
\item  A dual-branch architecture, using one branch for camera feature extraction from RGB image and a separate branch for LiDAR feature extraction from projected depth. Features are then merged at a later stage before being matched by a third backbone and finally regressed by the head;
\item Multi-scale refinement. By training several networks on different de-calibration ranges, then passing the input successively through these different networks at the time of inference to refine the output from a gross estimation to a more accurate value;
\item A well-thought split of Kitti~\cite{geigerVisionMeetsRobotics2013}, with a challenging testing set.
\end{itemize}

Another significant contribution to the field, CalibNet~\cite{iyerCalibNetGeometricallySupervised2018} introduced the use of new losses comparing the ground truth point cloud with the re-calibrated point cloud to take spatial information into account. 
NetCalib~\cite{wuNetCalibNovelApproach2021}\cite{wuThisWaySensors2021} introduces the idea of computing a depth map from stereo cameras to match with the LiDAR-based depth map. Using the same modalities achieves more accurate results.
Finally, one of the most recent publications, LCCNet~\cite{lvLCCNetLiDARCamera2021}, achieves the best results yet. It uses the common architecture previously introduced, as well as scale refinement and iterative refinement, and relies on a cost-volume~\cite{sunPWCNetCNNsOptical2018} as its backbone. To our knowledge, no prior work to UniCal has yet used  Transformers-based networks for the task of camera-to-LiDAR calibration. However, some works have used Transformer-based vision backbones for tasks relying on data from LiDAR and camera, such as~\cite{baiTransFusionRobustLiDARCamera2022}, which relies on a pre-calibrated system to perform fusion-based object detection.

These previous works achieve accurate calibration but leave room for improvement in accuracy, performance, and reliability in unstructured environments. The networks used are typically large as they follow a dual-branch architecture which requires two separate feature encoders for the camera branch and the LiDAR branch. As previous publications evaluate their results on different splits of the Kitti dataset, we show in our experiments (see Table~\ref{datasets} and Table~\ref{results}) that the results can vary significantly depending on the dataset split chosen. We thus also evaluate UniCal on the different splits used in previous references to compare the results faithfully.

\section{Method}

\begin{figure*}[t] 
    \centering
    \includeinkscape[width=0.85\textwidth]{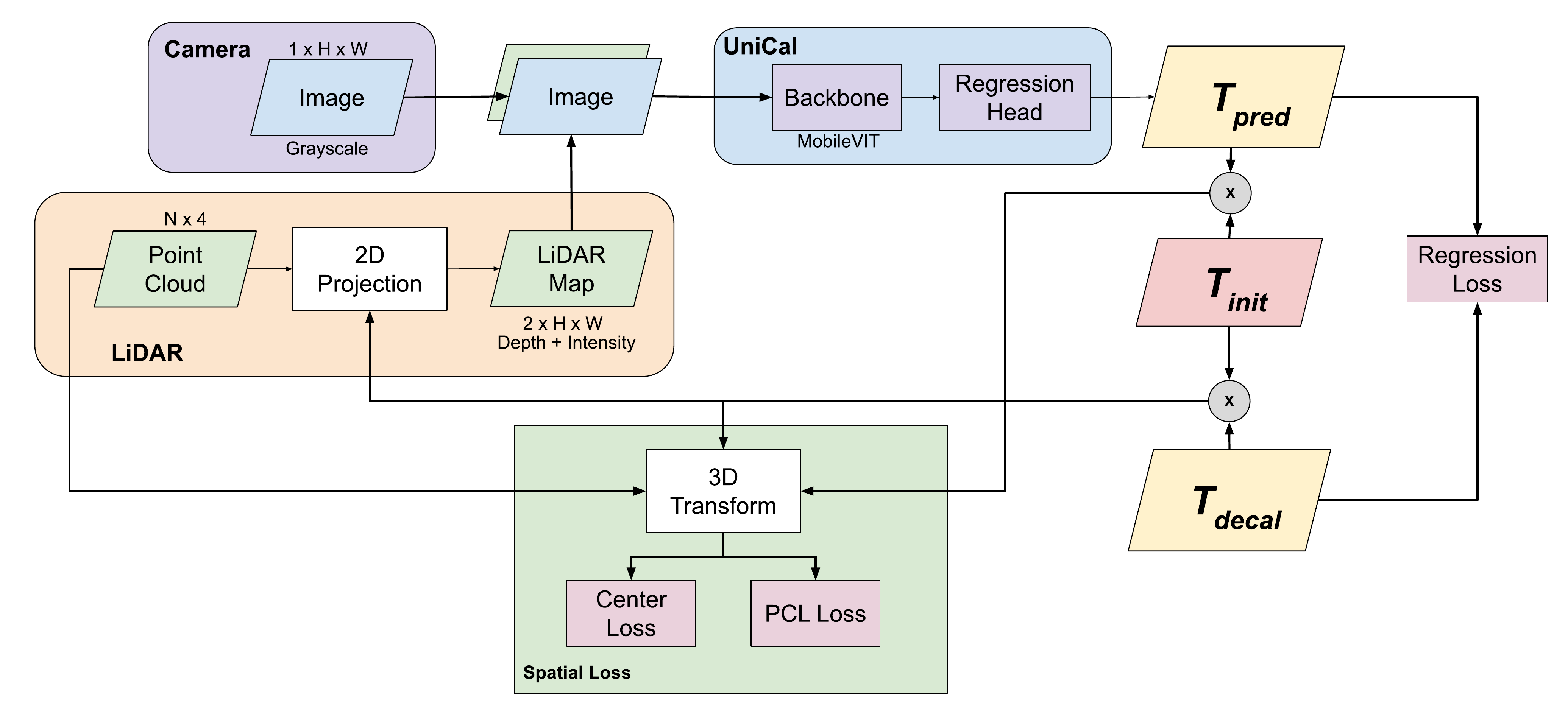}
    \caption{UniCal macro-architecture. The network receives data from the camera and LiDAR, merges them in a unique branch, and processes them through its backbone. In our case, we chose MobileViT~\cite{mehtaMobileViTLightweightGeneralpurpose2022}, but another model could work as well.}
    \label{fig:arch}
\end{figure*}

\subsection{Data Representation}
To perform the calibration task, we focus on the space of autonomous vehicles consisting of driving scenes. Each scene sample is a synchronized capture of a camera image and a LiDAR point cloud. For training, we need the corresponding sensor input pair and the ``ground truth" extrinsic calibration between them. Note that it is impossible to obtain the perfect extrinsic calibration outside of simulation. We assume that the intrinsic parameters are already calibrated.  

\subsubsection{Sample generation} Our dataset consists of calibrated scenes, we generate de-calibrated samples to feed as input to the network by following the method introduced by RegNet~\cite{schneiderRegNetMultimodalSensor2017}. In short, we generate a random de-calibration on all 6 DoF transformation parameters according to a uniform distribution on a chosen de-calibration range. We choose the range $\pm 10cm$ for translation and $\pm 1^\circ$ for rotation which is an estimate of the levels of perturbation experienced by the sensors during vehicle operation. We note $T$ a transformation with three rotation parameters and three translation parameters. This de-calibration $T_{decal}$ is then applied to the ground truth $T_{gt}$ to get the de-calibrated initial transformation $T_{init}$, as explained in Eq.~\ref{equation:T_generate}. We then project the LiDAR points into the camera frame using these de-calibrated parameters. Using $T_{init}$ as the initial calibration value, UniCal estimates the transformation $T_{decal}$ as $T_{pred}$, such that applying $T_{pred}^{-1}$ to $T_{init}$ should result in the ground truth value $T_{gt}$ as described in Eq.~\ref{equation:T_solve}. 
\begin{equation}
\label{equation:T_generate}
T_{init} = T_{decal} T_{gt}
\end{equation}
\begin{equation}
T_{gt} \approx T_{pred}^{-1}  T_{init}
\label{equation:T_solve}
\end{equation}

\subsubsection{Unified representation} Unlike previous works in this field~\cite{schneiderRegNetMultimodalSensor2017}\cite{iyerCalibNetGeometricallySupervised2018}\cite{wuNetCalibNovelApproach2021}\cite{shiCalibRCNNCalibratingCamera2020}, UniCal uses a unified representation of all data sources as the input of the network. We then extract features directly from this representation. This unified representation is an N-channel pseudo-image. Each channel corresponds to a different input source from a sensor (at least one channel per sensor). An advantage of this representation is that it requires fewer parameters than a dual-branch architecture, hence UniCal would require less computational resources. Moreover, it also makes UniCal more modular and extensible. The model can easily be extended to experiment with new input sources, whether they are from a different sensor (to try to solve a different calibration task) or a processed input from the same sensor (for example adding edge extractions on the camera image). In this work, we use grayscale from the camera, and depth and intensity from the LiDAR: we feed to the network a pseudo-image with 3 channels.

\subsection{Network Architecture} \label{section:architecture}
We introduce a novel architecture, illustrated in Fig.~\ref{fig:arch}. It is created to be a lightweight, modular, and efficient baseline that can easily be extended. Our early unified data representation allows us to replace the dual-branch architecture used in other works with a novel single-branch architecture. UniCal directly learns to match the different modalities with a single backbone network as opposed to the dual-branch architecture found in~\cite{schneiderRegNetMultimodalSensor2017}~\cite{lvLCCNetLiDARCamera2021}~\cite{iyerCalibNetGeometricallySupervised2018}~\cite{wuNetCalibNovelApproach2021}~\ which requires three separate backbones for feature extraction and feature matching. This makes UniCal lighter in comparison. 
Moreover, the architecture of UniCal and its backbone is not customized for the input type. The use of our unified input representation means that changing the inputs does not require other changes in the backbone than its channel number. 

\subsubsection{Backbone}
Recently, Transformer-based architectures have been successful in bringing the benefits of self-attention mechanisms to vision tasks. However, to learn jointly on all modalities, convolutional operations are still required. We use MobileViT~\cite{mehtaMobileViTLightweightGeneralpurpose2022} as the backbone. It implements a convolutional operation in which the local matrix multiplication is replaced by a global operation through a stack of Transformer~\cite{vaswaniAttentionAllYou2017} layers. It combines advantages from both convolutional networks (such as spatial bias) and Transformers (self-attention). Moreover, it was designed to be lightweight and to run on embedded systems with constrained resources. As a result, UniCal counts only 5.7 million trainable parameters.  In comparison, a ResNet18~\cite{heDeepResidualLearning2015} backbone has around 11 million trainable parameters. A network using a ResNet18-based backbone in a 2-branch architecture, such as CalibNet~\cite{iyerCalibNetGeometricallySupervised2018}, we estimate could have up to 33 million parameters for its backbone alone.  
For our experiments, we use the MobileViT~\cite{mehtaMobileViTLightweightGeneralpurpose2022} implementation proposed by Hugging Face~\cite{wolfTransformersStateoftheArtNatural2020}.

\subsubsection{Head} We use a simple regression head represented by fully connected layers to regress the calibration parameters which are 3 translation parameters (x, y, z) and 3 rotation parameters (roll, pitch, yaw). Our regression head consists of a common first layer which then splits into two branches to separately regress the translation and rotation components. This head is similar to the one used in~\cite{lvLCCNetLiDARCamera2021}.

\subsection{Loss Functions}
\subsubsection{Regression loss}
UniCal is trained using supervised learning. We use Mean Square Error regression losses for rotation as in Eq.~\ref{equation:loss_reg_r} and translation as in Eq.~\ref{equation:loss_reg_t} to compare the prediction and the ground truth de-calibration. In Eq.~\ref{equation:loss_reg_r} and Eq.~\ref{equation:loss_reg_t}, $r$ and $t$ are respectively the rotation and translation parameters of the transformation. Both losses  in Eq.~\ref{equation:loss_reg_r} and Eq.~\ref{equation:loss_reg_t} must then be averaged for the batch.

\begin{equation}\label{equation:loss_reg_r} \mathcal{L}_r =
    \left( r_{gt}  - r_{pred}\right)^{2}
\end{equation} 
\begin{equation}\label{equation:loss_reg_t} \mathcal{L}_t =
    \left( t_{gt}  - t_{pred}\right)^{2}
\end{equation}

\subsubsection{Spatial losses.}
Similarly to ~\cite{iyerCalibNetGeometricallySupervised2018}\cite{wuThisWaySensors2021}, we use spatially-aware losses to improve convergence during training. We use two such losses to compare the correct point cloud and the point cloud after re-calibration:
\begin{itemize}
    \item \textbf{Center loss}: the distance between the center of those two point clouds as in Eq.~\ref{equation:loss_center} where $C_{pcl}$ is the center of the point cloud. This loss must then be averaged for the batch;
    \begin{equation} \label{equation:loss_center}\mathcal{L}_{C} =
            \left( T_{gt} C_{pcl} - 
            T_{pred}^{-1}
            T_{init}    C_{pcl}\right)^{2}
    \end{equation}

    \item \textbf{Point cloud loss}: the distance between the corresponding points in those two point clouds (there is no need for matching as the data remains ordered) as in Eq.~\ref{equation:loss_pcl} where $p_k$ is a point from the point cloud with index $k$, and $K$ the number of points in the point cloud. This loss must then be averaged for the batch.
    \begin{equation}       \label{equation:loss_pcl}
        \mathcal{L}_{pcl} =
       \frac{1}{K}\sum_{1}^{K}\left( T_{gt}p_k - T_{pred}^{-1}T_{init}   p_k\right)^{2} 
    \end{equation}

\end{itemize}

\subsection{Self-Attention}
\begin{figure}[t]
     \centering
     \subfloat[Kitti image]{
         \includegraphics[width=0.45\columnwidth]{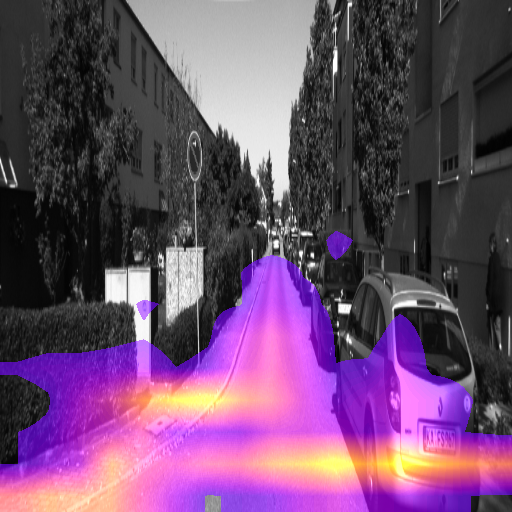}}
     \hfill
     \subfloat[Motional image]{
         \includegraphics[width=0.45\columnwidth]{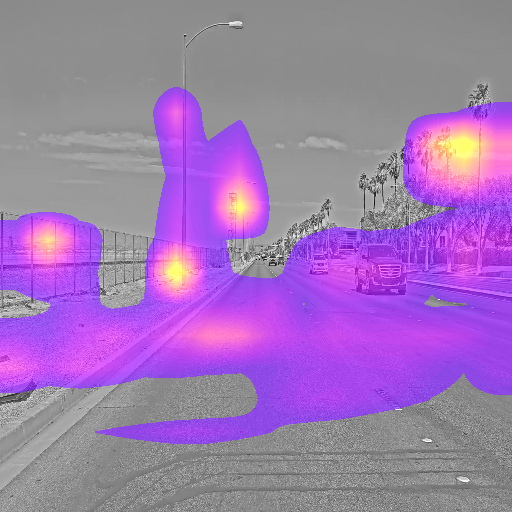}}
    \caption{Attention heatmaps overlayed on images from our datasets. Images are converted to grayscale and resized.}
    \label{fig:heatmaps}
\end{figure}

As detailed in~\ref{section:architecture}, we selected MobileViT~\cite{mehtaMobileViTLightweightGeneralpurpose2022} as our backbone because it is adapted to embedded applications and brings benefits of both convolutional networks and Transformers. Self-attention mechanisms used in Transformers allow UniCal to give more weight to features deemed more relevant by the network. Attention also brings some form of explainability as it can be displayed as a heatmap for us to see the areas the network found most relevant to solve the calibration task. The heatmaps generated by UniCal during testing are illustrated in Fig.~\ref{fig:heatmaps}.

We observed that UniCal weights its attention in different zones depending on the dataset on which it was trained. On Kitti, it tends to give more attention weight to roads and cars. On Motional drivelog data, it tends to highlight various salient objects in the image, especially objects that offer clear lines visible with both sensors. Those zones with high attention closely resemble those that humans experimented with calibrating would look at to spot de-calibration.
We suppose that the network can often rely on elements found on the road with Kitti data as it is less challenging than the Motional data and only uses the front camera where the road is always visible.

\section{Experiments}
\subsection{Dataset}
\subsubsection{Kitti} We decided to work mostly with Kitti~\cite{geigerVisionMeetsRobotics2013}, a reference dataset on the autonomous driving scene. Kitti has been used in all relevant works on this topic. It provides enough samples and an accurate calibration ground truth. Using the same dataset as previous works will help us compare our results. However, those previous publications used different splits of the Kitti dataset. We present those splits in Table~\ref{datasets} and will refer to them as $\alpha$, $\beta$, $\gamma$, and $\delta$.

\begin{table*}[h]
\caption{Kitti~\cite{geigerVisionMeetsRobotics2013} splits used in state-of-the-art works} \vspace{-4mm}
\label{datasets}
\begin{center}
\begin{tabular}{cccccc} \hline
\textbf{Split} & \textbf{Publication} & \textbf{Dataset} & \multicolumn{1}{c}{\textbf{Training}} & \textbf{Validation} & \textbf{Testing} \\ \hline \hline
$\alpha$ & RegNet~\cite{schneiderRegNetMultimodalSensor2017} & \begin{tabular}[c]{@{}c@{}}Raw\end{tabular} & \multicolumn{1}{c}{\begin{tabular}[c]{@{}c@{}}26\_09: all logs not used in other sets\end{tabular}} & \begin{tabular}[c]{@{}c@{}}26\_09: 5, 70\end{tabular} & \begin{tabular}[c]{@{}c@{}}30\_09: 28\end{tabular} \\ 
$\beta$ & NetCalib~\cite{wuThisWaySensors2021} & \begin{tabular}[c]{@{}c@{}} Raw\end{tabular} & \multicolumn{1}{c}{\begin{tabular}[c]{@{}c@{}}26\_09: all logs not used in other sets\end{tabular}} & \begin{tabular}[c]{@{}c@{}}26\_09: 13, 20, 79\end{tabular} & \begin{tabular}[c]{@{}c@{}}26\_09: 5, 70\end{tabular} \\ 
$\gamma$ & CalibNet~\cite{iyerCalibNetGeometricallySupervised2018} & \begin{tabular}[c]{@{}c@{}} Raw\end{tabular} & 26\_09: unspecified logs  & 26\_09: unspecified logs  & \begin{tabular}[c]{@{}c@{}}26\_09, 30\_09: unspecified logs\end{tabular} \\ 
$\delta$ & LCCNet~\cite{lvLCCNetLiDARCamera2021} & \begin{tabular}[c]{@{}c@{}} Odometry\end{tabular} & logs 1~--~20: unspecified distribution  & logs 1~--~20: unspecified distribution  & log 0 \\ \hline
\end{tabular}%
\end{center}
\vspace{1mm}
\textit{N.B.: In each split, sets are mutually exclusive. While logs from a same day can be used in different sets, each log can only belong to one set. Here "26\_09: 5" identifies log 5 of day 26/09.}
\vspace{-1mm}
\end{table*}

From the splits presented in Table~\ref{datasets},  we choose $\alpha$ as our reference. Proposed by RegNet~\cite{schneiderRegNetMultimodalSensor2017}, it is the first Kitti split introduced for this task, and as shown by our results in Table~\ref{results} it remains the most challenging. This is because it uses samples from separate days for training/validation and testing, with different camera intrinsics. In comparison, splits $\beta$, $\gamma$, and $\delta$ used in other publications include in their testing sets samples recorded the same day and with the same camera intrinsics as their training sets. $\delta$ even uses spatially redundant samples (some scenes are captured in the same location in training and testing sets) as acknowledged in~\cite{lvLCCNetLiDARCamera2021}. The split and dataset choice are not trivial as results in Table~\ref{results} show that the same network (here UniCal) trained and tested on different splits will perform differently. The splits with more similarities between the training and testing sets will get seemingly better results, whether those similarities are mostly in the camera intrinsics as in $\beta$, or even in the location where the samples were collected, as in $\delta$.

\subsubsection{Motional drive log data}\label{motional_data} To further assess our method, we also generated our own dataset collected from 89 autonomous vehicle driving logs across 4 cities (Las Vegas, Santa Monica, Pittsburgh, Boston) consisting of 28 unique vehicles. We use data from one main LiDAR and 8 different cameras surrounding the vehicle. We split our dataset into 21995 training data, 3299 validation data, and 2173 testing data. Our training and validation data consists of logs taken from only Las Vegas and Santa Monica whereas our testing data has logs from all 4 locations. We also made sure our testing data has no overlapping vehicles with the training set.

\subsubsection{Data augmentation}
Our best results are obtained with soft augmentation parameters to fit real-life situations as much as possible. We can expect that the vehicle will always lie flat on the road and the ground will be approximately horizontal or rotated with a limited angle. We thus augmented with random rotations of up to 2$^\circ$, translation of 0.01$\%$ of the image dimensions.

\subsection{Results}
\subsubsection{Kitti} We compare in Table~\ref{results} the results obtained on the different splits presented in Table~\ref{datasets}. The models compared have been trained to solve the task on different de-calibration ranges. However, they rely on \textit{multi-scale refinement} to bring the final task to the same de-calibration range: a cascade of networks are trained on different ranges of de-calibration with each step refining the parameters to a range suitable for the next network. The final network, trained on the smallest de-calibration range, receives parameters that are already in its range. This last network can thus be considered independently as operations happening beforehand are transparent to it. It is then possible to compare results from this final network with results from a network trained on the same de-calibration range.

\begin{table}[b]
\caption{Results and comparison with other works on Kitti}
\label{results}
\begin{center}
\begin{tabular}{cccccc} \hline
\multirow{2}{*}{\textbf{Model}} & \multirow{2}{*}{\textbf{Split}} & \multicolumn{2}{c}{\textbf{Rotation ($^{\circ}$)}} & \multicolumn{2}{c}{\textbf{Translation (cm)}} \\
 &  & \multicolumn{1}{c}{\textit{\textbf{MAE}}\footnotemark[1]} & \textit{\textbf{STD}\footnotemark[2]} & \multicolumn{1}{c}{\textit{\textbf{MAE}}} & \textit{\textbf{STD}} \\ \hline\hline
UniCal (ours) & \multirow{2}{*}{$\alpha$} & \multicolumn{1}{c}{\textbf{0.04}} & \textbf{0.03} & \multicolumn{1}{c}{\textbf{0.89}} & \textbf{0.85} \\ 
RegNet~\cite{schneiderRegNetMultimodalSensor2017} &  & \multicolumn{1}{c}{0.28} & - & \multicolumn{1}{c}{6} & - \\ \hline
UniCal (ours) & \multirow{2}{*}{$\beta$} & \multicolumn{1}{c}{\textbf{0.03}} & \textbf{0.03} & \multicolumn{1}{c}{\textbf{0.33}} & \textbf{0.30} \\ 
NetCalib~\cite{wuThisWaySensors2021} &  & \multicolumn{1}{c}{0.11} & 0.11 & \multicolumn{1}{c}{1.13} & 0.97 \\ \hline
CalibNet~\cite{iyerCalibNetGeometricallySupervised2018} & $\gamma$ & \multicolumn{1}{c}{0.41} & - & \multicolumn{1}{c}{4.34} & - \\\hline
UniCal (ours) & \multirow{2}{*}{$\delta$} & \multicolumn{1}{c}{0.04} & 0.03 & \multicolumn{1}{c}{0.80} & 0.80 \\ 
LCCNet~\cite{lvLCCNetLiDARCamera2021} & & \multicolumn{1}{c}{\textbf{0.017}} & - & \multicolumn{1}{c}{\textbf{0.30}} & - \\ \hline
\end{tabular}
\end{center}
\end{table}

On Kitti, we observe that UniCal outperforms most networks from the state of the art, notably RegNet~\cite{schneiderRegNetMultimodalSensor2017}, CalibNet~\cite{iyerCalibNetGeometricallySupervised2018}, and NetCalib~\cite{wuThisWaySensors2021}. However, as expected it comes short of LCCNet~\cite{lvLCCNetLiDARCamera2021}, which relies on multiple scenes for each prediction to refine its results. By contrast, we chose to estimate the calibration parameters with a single shot. 

\subsubsection{Motional proprietary data} \label{motional_data_sss}
We experimented on the proprietary data described in~\ref{motional_data}. This dataset is challenging for a calibration network: the data is from multiple vehicles, captured in different conditions and locations, and intrinsic calibration, as well as extrinsic ground truth, can be imperfect. Most importantly, we made the task more difficult by using data from all available cameras on the vehicle compared to only one camera in our Kitti experiments and state-of-the-art works. UniCal is to our knowledge the first model that has been shown able to calibrate a lidar with multiple cameras. Those cameras have wildly different points of view and positions on the vehicle (front, back, sides, etc.). We achieved a mean average error of $0.13^\circ$ on rotation and $1.9 cm$ on translation. To better understand those results, this is still at least twice more accurate than the $0.28^{\circ}$ and $6cm$ achieved by RegNet~\cite{schneiderRegNetMultimodalSensor2017} on Kitti for a single camera, despite our task and dataset being more challenging.
\footnotetext[1]{MAE: Mean Average Error}
\footnotetext[2]{STD: Standard Deviation}

\subsection{Execution time}

By profiling UniCal on an NVIDIA GeForce RTX 2070 SUPER, we get an inference time of around $11.67ms$ for a batch size of $1$. This means UniCal can perform calibration about 85 times per second, which is enough for real-time applications. The calibration can thus be performed in milliseconds while the vehicle is operating. A traditional vision method such as~\cite{yuanPixellevelExtrinsicSelf2021} is much slower, with data acquisition taking about $20s$~\cite{yuanPixellevelExtrinsicSelf2021} during which the vehicle must not move, and the processing pipeline takes another $60s$~\cite{yuanPixellevelExtrinsicSelf2021}. Compared  to other deep learning-based methods, it is difficult to establish any significant inference time difference. \cite{wuNetCalibNovelApproach2021} reports an inference time of $16ms$, while \cite{wuThisWaySensors2021} reports $4.7ms$ for a similar network. Moreover, measurement methods and equipment can vary, making it difficult to compare results. We can thus infer that deep learning methods should all have a relatively low inference time, measured in milliseconds compared to traditional vision methods for which execution time is most often in the scale of seconds such as \cite{yuanPixellevelExtrinsicSelf2021}. 

\begin{table}[tbhp]
\caption{Ablation study results on Kitti~\cite{geigerVisionMeetsRobotics2013}}
\vspace{-4mm}
\label{ablation}
\begin{center}
\resizebox{\columnwidth}{!}{%
\begin{tabular}{cccc|cccc} \hline
\multicolumn{4}{c|}{\textbf{Experiment}}                                                                                                            & \multicolumn{2}{c}{\textbf{Rotation (°)}}       & \multicolumn{2}{c}{\textbf{Translation (cm)}}   \\ 
\multicolumn{1}{c}{\textbf{Augmentation}} & \multicolumn{1}{c}{\textbf{Intensity}} & \multicolumn{1}{c}{\textbf{Attention}} & \textbf{Grayscale} & \multicolumn{1}{c}{\textbf{MAE\footnotemark[1]}} & \multicolumn{1}{c}{\textbf{STD\footnotemark[2]}} & \multicolumn{1}{c}{\textbf{MAE}} & \textbf{STD} \\  \hline \hline
\multicolumn{1}{c}{}                      & \multicolumn{1}{c}{\checkmark}          & \multicolumn{1}{c}{\checkmark}          & \checkmark          & \multicolumn{1}{c}{0.04}         & 0.04         & \multicolumn{1}{c}{1.06}         & 1.16         \\  
\multicolumn{1}{c}{\checkmark}             & \multicolumn{1}{c}{\checkmark}          & \multicolumn{1}{c}{\checkmark}          & \checkmark          & \multicolumn{1}{c}{0.04}         & 0.03         & \multicolumn{1}{c}{0.89}         & 0.85         \\ 
\multicolumn{1}{c}{}                      & \multicolumn{1}{c}{}                   & \multicolumn{1}{c}{\checkmark}          & \checkmark          & \multicolumn{1}{c}{0.04}         & 0.04         & \multicolumn{1}{c}{0.96}         & 0.96         \\ 
\multicolumn{1}{c}{}                      & \multicolumn{1}{c}{\checkmark}          & \multicolumn{1}{c}{}                   & \checkmark          & \multicolumn{1}{c}{0.05}         & 0.06         & \multicolumn{1}{c}{1.37}         & 1.24         \\ 
\multicolumn{1}{c}{}                      & \multicolumn{1}{c}{\checkmark}          & \multicolumn{1}{c}{\checkmark}          &                    & \multicolumn{1}{c}{0.05}         & 0.05         & \multicolumn{1}{c}{1.26}         & 1.13         \\ \hline
\end{tabular}%
}
\end{center}
\end{table}
\subsection{Ablation study} \label{subsec_ablation}
We ran additional experiments to study the influence of different choices on our model. The results are reported in Table~\ref{ablation}. Those experiments were conducted on Kitti, using the split $\alpha$ presented in Table~\ref{datasets}.

\subsubsection{Augmentation}
 Results in Table~\ref{ablation} show that applying a soft data augmentation improves results in rotation and translation. We can infer that data augmentation helps the network to learn more features and generalize better.
\\

\subsubsection{LiDAR intensity}
\begin{figure}[htbp]
    \centering
    \includegraphics[width=\columnwidth]{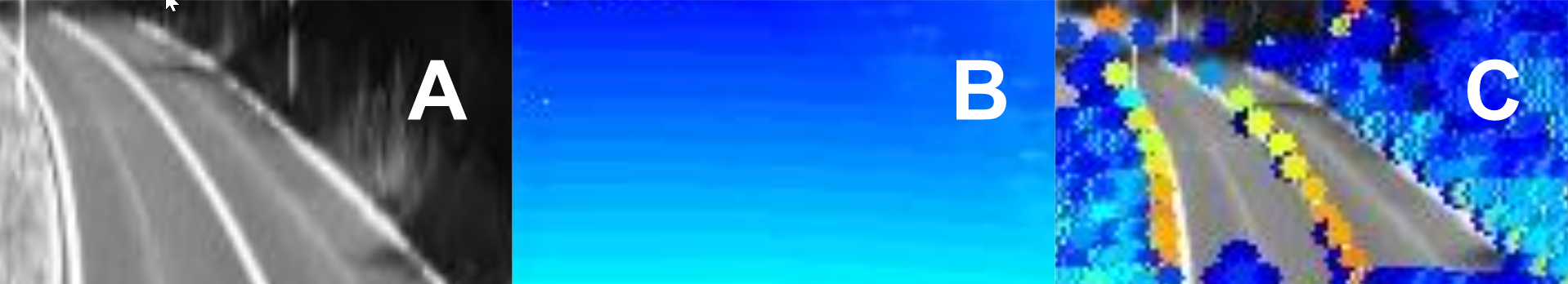}
    \caption{Crop from a Kitti image. The lines from the road marking can be seen on the grayscale image (\textbf{A}) and the intensity projection (\textbf{C}), but not on the depth projection (\textbf{B}).}
    \label{fig:IvR}
\end{figure}

Intuitively, adding more information should help the network learn new features. Some datasets, like Kitti, provide intensity information from the LiDAR sensor. As shown in Fig.~\ref{fig:IvR}, intensity can be helpful in perceiving two-dimensional visual features and patterns that are not perceived in the depth map. This is the case in Fig.~\ref{fig:IvR} where road surface markings, which are visible in the intensity map, are not visible in the depth map since they are two-dimensional. However, our experiments in~\ref{ablation} showed that this information is not always helpful for our task. This is probably because it makes visible to the network some elements that are noisy or not meaningful for our task.

\subsubsection{Self-attention}
To try and determine the contribution of attention mechanisms in UniCal, we ran an experiment in which we replaced the MobileViT~\cite{mehtaMobileViTLightweightGeneralpurpose2022} backbone with a ResNet~\cite{heDeepResidualLearning2015} backbone. ResNet is a popular CNN architecture for vision tasks that does not use attention mechanisms. As expected, the results in Table~\ref{ablation} show about 25\% higher Mean Average Error (MAE) and Standard Deviation (STD) on rotation and translation parameters compared to our reference. This shows that the selected MobileViT backbone outperforms traditional architectures in UniCal. Considering that relevant features are correctly found by attention mechanisms in our heatmap visualizations in~\ref{fig:heatmaps}, this is additional evidence of the relevance of attention for this task.
However, results obtained with ResNet are still satisfactory and close to the state-of-the-art, showing that the unified representation will work with different network backbones. 

\subsection{Calibration Validation}

\begin{table}[h]
\caption{calibration validation results on Motional data} \vspace{-4mm}
\label{calval}
\begin{center}
\begin{tabular}{c|c|cccc} \hline
\textbf{Model}     & \textbf{TL\footnotemark[1]}                                                     & \textbf{Accuracy} & \textbf{F1} & \textbf{Precision} & \textbf{Recall} \\ \hline \hline
RegNet-based &                                                                 & 0.52            & 0           & 0                  & 0               \\ 
UniVal &                                                              & 0.97            & 0.97      & 0.98             & 0.95          \\ 
UniVal & \checkmark & \textbf{0.98}   & \textbf{0.98} & \textbf{0.99}    & \textbf{0.97}          \\ \hline
\end{tabular}
\end{center}
\end{table}
\footnotetext[1]{TL: Transfer Learning}

\addtolength{\textheight}{-9cm}   %

Besides calibration, we also looked into the task of calibration validation. Given a corresponding pair of camera image and LiDAR point cloud, the task is to detect if the calibration is correct (within an acceptable tolerance range). We pose this problem as a binary classification task. A positive classification would be a correct calibration within the tolerance range, and a negative classification indicates a de-calibration. We perturb the calibration by up to $1^\circ$ on each rotation axis and up to $10 cm$ on each translation axis. 

Initially, we switch out our network heads for classification heads and trained on Kitti. We aptly named this new network UniVal. From our results in Table~\ref{calval}, we found that the RegNet model severely overfits during training while our UniVal model fairs significantly better. We then tried a transfer learning method that involves pre-training the network weights on the calibration task, freezing the network layers, and fine-tuning with a validation (classification) head. The motivation is that the backbone weights trained on calibration should be expressive enough to be used in the validation task. Our results show that the performance using this method improves from simply training UniVal without transfer learning.

\section{Conclusion}
In this work, we showed that self-attention-based vision can be leveraged to improve learning on driving scenes and proposed a single-branch architecture that outperforms the standard dual-branch architecture introduced by RegNet~\cite{schneiderRegNetMultimodalSensor2017}. Besides that, we also show that we are able to use transfer learning from the calibration task to outperform regular training on the calibration validation task without retraining the calibration network weights. For future work, we propose the following:
\\

\subsubsection{Refinement and re-calibration ranges.} We focused on single-shot calibration for the small ranges of de-calibration most often occurring while operating an autonomous vehicle. Moreover, we aimed to develop a new architecture that could be used as a baseline and extended in future works. However, one could further improve results by adding some form of refinements such as the multi-frame iteration introduced in RegNet~\cite{schneiderRegNetMultimodalSensor2017} as \textit{temporal filtering} or the LSTM-based refinement introduced by CalibRCNN~\cite{shiCalibRCNNCalibratingCamera2020}. Similarly, we could handle large de-calibration by using scale-iterative refinement as in RegNet~\cite{schneiderRegNetMultimodalSensor2017} and others.
\\

\subsubsection{Extension to other modalities.} As a future work, we believe that our model can be extended to other tasks by simply modifying the modality of input used in the unified representation. This would enable calibration of other modalities such as Camera-to-Camera calibration or LiDAR-to-LiDAR calibration. We could also try to regress camera intrinsic parameters (focal length, principal point, distortion coefficients) to perform intrinsic calibration as well.

\section*{Acknowledgment}

We would like to thank our colleagues Huy Nguyen, Venice Erin Liong, Billy Saputra for their insightful advice and help in reviewing this work.

\bibliographystyle{resources/bibliography/IEEEtran.bst}
\bibliography{resources/bibliography/biblio.bib}

\end{document}